\documentclass[11pt]{article} 
\usepackage{rldm,palatino}

\usepackage{subcaption}
\usepackage{graphicx,wrapfig,adjustbox,lipsum}
\usepackage[inline]{enumitem}
\usepackage{amsthm}
\usepackage[ruled,noend]{algorithm2e}

\title{World Value Functions: Knowledge Representation for Multitask Reinforcement Learning}

\author{
Geraud Nangue Tasse, Steven James, Benjamin Rosman \\
School of Computer Science and Applied Mathematics\\
University of the Witwatersrand\\
Johannesburg, South Africa \\
\texttt{geraud.nanguetasse1@students.wits.ac.za, \{steven.james,benjamin.rosman1\}@wits.ac.za} \\
}

\newtheorem{experiment}{Experiment}[section]
\newtheorem{definition}{Definition}
\newtheorem{theorem}{Theorem}

\newlength{\strutheight}
\usepackage{amsmath,amsfonts,bm}

\def\eqref#1{equation~\ref{#1}}

\def\1{\bm{1}}

\DeclareMathAlphabet{\mathsfit}{\encodingdefault}{\sfdefault}{m}{sl}
\SetMathAlphabet{\mathsfit}{bold}{\encodingdefault}{\sfdefault}{bx}{n}

\newcommand{\E}{\mathbb{E}}

\DeclareMathOperator*{\argmax}{arg\,max}

\usepackage{calc}
\usepackage{amsmath}
\usepackage{mathtools}
\usepackage{mathrsfs}
\usepackage{array}
\newcolumntype{P}[1]{>{\centering\arraybackslash}p{#1}}

\makeatletter
\providecommand*{\barvee}{%
  \mathbin{%
    \mathpalette\@barvee{}%
  }%
}
\newcommand*{\@barvee}[2]{%
  \sbox0{$#1\veebar\m@th$}%
  \sbox2{%
    \hbox to \wd0{%
      \hss
      \resizebox{1.05\wd0}{\height}{$#1-\m@th$}%
      \hss
    }%
  }%
  \sbox4{%
    \resizebox{\wd0}{.7\ht0}{$#1\vee\m@th$}%
  }%
  \sbox6{$#1\vcenter{}$}
  \ht2=\ht6 %
  \vbox to \ht0{%
    \copy2 %
    \vss
    \copy4 %
  }%
}
\makeatother

\newcommand{\state}{\mathcal{S}}
\newcommand{\action}{\mathcal{A}}
\newcommand{\dynamics}{P}
\newcommand{\reward}{R}
\newcommand{\rmax}{\reward_{\text{MAX}}}
\newcommand{\rmin}{\reward_{\text{MIN}}}
\newcommand{\goals}{\mathcal{G}}
\renewcommand{\v}{V}
\newcommand{\q}{Q}
\newcommand{\vpi}{V^\pi}
\newcommand{\qpi}{Q^\pi}
\newcommand{\pistar}{\pi^*}

\newcommand{\vstar}{V^*}
\newcommand{\qstar}{Q^*}

\newcommand{\tasks}{\mathcal{M}}

\usepackage{mathtools}

\newcommand{\mbig}{\tasks_{SUP}}
\newcommand{\msmall}{\tasks_{INF}}

\newcommand{\rbar}{\bar{\reward}}
\newcommand{\rbarmin}{\rbar_{\text{MIN}}}
\newcommand{\pibar}{\bar{\pi}}
\newcommand{\pibarstar}{\bar{\pi}^*}

\newcommand{\vbarpi}{\bar{\v}^{\pibar}}
\newcommand{\vstarbar}{\bar{\v}^*}
\newcommand{\vstarbara}{\Tilde{\bar{\v}}}
\newcommand{\qbar}{\bar{\q}}
\newcommand{\qbarpi}{\bar{\q}^{\pibar}}
\newcommand{\qstarbar}{\bar{\q}^{*}}

\newcommand{\qstarbarbig}{\qstarbar_{SUP}}
\newcommand{\qstarbarsmall}{\qstarbar_{INF}}

\newcommand{\goalq}{\bar{\mathcal{\q}}^*}

\newcommand{\gstar}{G^*_{s, g, a}}

\begin{document}

\maketitle

\begin{abstract}

An open problem in artificial intelligence is how to learn and represent knowledge that is sufficient for a general agent that needs to solve multiple tasks in a given world.
In this work we propose world value functions (WVFs), which are a type of general value function with mastery of the world---they represent not only how to solve a given task, but also how to solve any other goal-reaching task.
To achieve this, we equip the agent with an internal goal space defined as all the world states where it experiences a terminal transition---a task outcome.
The agent can then modify task rewards to define its own reward function, which provably drives it to learn how to achieve all achievable internal goals, and the value of doing so in the current task.
We demonstrate a number of benefits of WVFs. 
When the agent's internal goal space is the entire state space, we demonstrate that the transition function can be inferred from the learned WVF, which allows the agent to plan using learned value functions.
Additionally, we show that for tasks in the same world, a pretrained agent that has learned any WVF can then infer the policy and value function for any new task directly from its rewards.   
Finally, an important property for long-lived agents is the ability to reuse existing knowledge to solve new tasks. Using WVFs as the knowledge representation for learned tasks, we show that an agent is able to solve their logical combination zero-shot, resulting in a combinatorially increasing number of skills throughout their lifetime.


\end{abstract}



\startmain 

\section{Introduction}
\label{sec:intro}

The ultimate goal of artificial intelligence is to create general agents capable of solving multiple tasks in the world in the same way that humans are able to.
To achieve this, we need a general decision-making framework for agents that interact with a world to solve tasks, and a sufficiently general knowledge representation for what those agents learn.  
Reinforcement learning (RL) \cite{sutton1998introduction} is one such framework that has made several major breakthroughs in recent years, ranging from robotics to board games, but these agents typically are narrowly designed to solve only a single task.

In RL, tasks are specified through a reward function from which the agent receives feedback. 
Most commonly, an agent represents its knowledge in the form of a value function, which represents the sum of future rewards it expects to receive.
However, since the value function is directly tied to one single reward function (and hence task), it is definitionally insufficient for constructing agents capable of solving a wide range of tasks.
We seek to overcome this limitation by proposing \textit{world value functions} (WVFs), a knowledge representation that encodes how to solve not just the current task, but any other task, in the world. 
In the literature, agents with such abilities are said to possess \textit{mastery} \cite{veeriah2018many}. 
Importantly, WVFs are a form of general value function \cite{sutton11} that can be learned from a single stream of experience like standard value functions---no additional information need be provided to the agent.  

In this work, we define the WVF and demonstrate how it can be learned using standard RL algorithms. 
We then demonstrate several advantages of learning such a representation. 
In particular, we show that (a) WVFs implicitly encode the dynamics of the world and can be used for model-based RL; (b) having learned a WVF, any new task that an agent encounters can be solved by estimating its reward function, reducing the problem to a supervised learning one; and (c) WVFs can be \textit{composed} with Boolean operators to produce novel and human-interpretable behaviours.


\section{World Value Functions} \label{sec:wvfs}

We model the agent's interaction with the world as a Markov Decision Process (MDP) $(\mathcal{\state}, \action, \dynamics, \reward)$, where 
\begin{enumerate*}[label=(\roman*)]
  \item $\state$ is the state space,
  \item $\action$ is the action space,
  \item $\dynamics(s, a, s')$ are the transition dynamics of the world, and
  \item $\reward$ is a reward function bounded by $[\rmin,\rmax]$, representing the task the agent needs to solve.
\end{enumerate*}
Note that in this work, we focus on environments with \textit{deterministic} dynamics, but put no restrictions on their complexity.

The agent's aim is to compute a \textit{policy} $\pi$ from $\state$ to $\action$ that optimally solves a given task.
This is often achieved through a value function that represents the expected return obtained under $\pi$ starting from state $s$: $\vpi(s) = \E^\pi \left[ \sum_{t=0}^{\infty} r(s_t, a_t, s_{t+1}) \right]$.
Similarly, a $Q$-value function $\qpi(s, a)$ represents the expected return obtained by executing $a$ from $s$, and thereafter following $\pi$.
The optimal $Q$-value function is given by $\qstar(s, a) = \max_\pi \qpi(s, a)$ for all states $s$ and actions $a$, and the optimal policy follows by acting greedily with respect to $\qstar$ at each state. 

\begin{wrapfigure}{r}{0.52\textwidth}
\vspace{-45pt}
\SetAlgoNoLine
\begin{algorithm}[H]
\label{alg:dq}
\DontPrintSemicolon
    \SetKwInOut{Initialise}{Initialise}
 \Initialise{ WVF $\qbar$, goal buffer $\goals$ \;}
\ForEach{episode}{
    Observe an initial state $s\in\state$ and sample a goal $g\in\goals$\; 
   \While{episode is not done}{
    $a \leftarrow 
    \begin{cases}
    \argmax\limits_{a \in \action} \qbar(s, g, a) & \mbox{probability  } 1 - \varepsilon  \\
    \text{a random action} & \mbox{probability } \varepsilon 
    \end{cases}$ \;
   Take action $a$, observe reward $r$ and next state $s^\prime$ \;
   \textbf{if} \textit{$s^\prime$ is absorbing} \textbf{then}  $\goals \leftarrow \goals \cup \{s\}$ \;
   \For{$g^\prime\in\goals$}{
    $\bar{r} \leftarrow \rbarmin$ \textbf{if} $g^\prime \neq s$ and $s \in \goals$ \textbf{else} $r$ \;
    $\qbar(s, g^\prime, a) \xleftarrow[]{\alpha} \left[ \bar{r} + \max\limits_{a^\prime} \qbar(s^\prime, g^\prime, a^\prime) \right] - \qbar(s, g^\prime, a)$\;
    }
    $s \leftarrow s^\prime$
   }
 }
 \caption{Q-learning for WVFs}
 \label{algo:qlearn}
\end{algorithm}
\vspace{-10pt}
\end{wrapfigure}

\subsection{A New Form of General Value Function}

We now introduce \textit{world value functions} (WVFs), which provably encode how to reach all achievable goals in a world with arbitrary task rewards.
We first define the internal goal space $\goals$ of the agent as all states where it experiences a terminal transition.
%
%
Importantly, the goal the agent wishes to achieve is not specified by the world, but rather chosen by itself.
In order to simultaneously learn how to achieve the overall task goal and the agent's internal goals, the agent can define its own reward 
function $\rbar$, which extends the task rewards $\reward$ to simply penalise itself for achieving goals it did not intend to:
\[
\rbar(s, g, a, s^\prime) \coloneqq \begin{cases}
\rbarmin & \text{if } g \neq s \text{ and } s^\prime \text{ is absorbing }\\
\reward(s, a, s^\prime) &\text{otherwise},
\end{cases}
\]
where $\rbarmin$ is a large negative penalty that can be derived from the bounds of the reward function \cite{nangue2020boolean}.
We can think of the penalty $\rbarmin$ as adding one bit of information to the agent's rewards, which we will later prove is sufficient for the agent to learn how to achieve its internal goals and the value of achieving them in the current task.
The overall goal of the agent now is to compute a \textit{world policy} $\pibar: \state \times \goals \to Pr(\action)$ that optimally reaches its internal goal states. Given a world policy $\pibar$, the corresponding WVF is defined as follows: 
\[
 \qbarpi(s, g, a) \coloneqq \E_{s^\prime}^{\pibar} \left[ \rbar(s, g, a, s^\prime) + \vbarpi(s^\prime, g) \right],
\text{ where }
 \vbarpi(s, g) \coloneqq \E^{\pibar} \left[ \sum_{t=0}^{\infty} \rbar(s_t, g, a_t, s_{t+1}) \right].
\]

Since the WVF satisfies the Bellman equations, 
$\qstarbar(s,g,a)$ can be learned using any suitable RL algorithm like Q-learning (Algorithm~\ref{algo:qlearn}).
To show that a WVF does still encode knowledge of how to solve the task in which it was learned, we prove that the regular task reward function and value function can be recovered by simply maximising over goals (Theorem~\ref{thm:1}). The task policy can then be obtained as:
$
\pistar(s) \in \argmax_{a \in \action} \left( \max_{g \in \goals} \qstarbar(s, g, a) \right).
$
\begin{theorem}
Let $M=(\mathcal{\state}, \action, \dynamics, \reward)$ be a task with optimal Q-value function $\qstar$ and optimal world Q-value function $\qstarbar$. 
Then for all $(s, a)$ in $\state \times \action$, we have
\begin{enumerate*}[label=(\roman*)]
    \item $\reward(s, a) = \max\limits_{g \in \goals} \rbar(s, g, a)$, and
    \item $\qstar(s, a) = \max\limits_{g \in \goals} \qstarbar(s, g, a)$.
\end{enumerate*}
\label{thm:1}
\end{theorem}

Having established WVFs as a type of task-specific GVF, we prove in Theorem~\ref{thm:mastery} that they do indeed have mastery---that is, they learn how to reach all reachable goal states in the world. We formally define mastery as follows:

\begin{definition}
Let $\qstarbar$ be the optimal world $Q$-value function for a task $M$. Then $\qstarbar$ has mastery if for all  $g \in \goals$ reachable from $s \in \state \setminus \goals$, there exists an optimal world policy
\[
\pibarstar(s,g) \in \argmax\limits_{a \in \action}\qstarbar(s, g, a)
\text{ such that }
\pibarstar \in \argmax\limits_{\pibar} P^{\pibar}_{s}(s_{T} = g),
\]
where $P^{\pibar}_{s}(s_{T} = g)$ is the probability of reaching $g$ from $s$ under a policy $\pibar$.
\label{def:mastery}
\end{definition}

\begin{theorem}
\label{theorem:2}
Let $\qstarbar$ be the optimal world $Q$-value function for a task $M$. Then $\qstarbar$ has mastery.
\label{thm:mastery}
\end{theorem}

Finally, if the agent's goal space is the state space ($\goals=\state$), then we can estimate the transition probabilities for each $s,a\in\state\times\action$ using only the reward function and optimal WVF. That is, $P(s,a,s')$ for all $s^\prime\in\state$ can be obtained by simply solving for them in the system of Bellman optimality equations given by each goal $ g\in\state$:  
$
\qstarbar(s,g,a) = \sum_{s' \in \state} p(s,a,s^\prime) \left[ \rbar(s,g,a,s') + \vstarbar(s',g) \right].
$
This shows that optimal WVFs implicitly encode the dynamics of the world.
In practice, if the learned WVF is sufficiently optimal in a neighbourhood $\mathcal{N}(s)$, then we only need to consider $s^\prime,g\in\mathcal{N}(s)\times\mathcal{N}(s)$ if the transition probabilities are known to be non-zero only in $\mathcal{N}(s)$, common in most domains. 

\begin{experiment}
\label{exp:WVF}

Consider a 4-rooms gridworld where an agent may be required to visit various goal positions. The agent can move in any of the four cardinal directions at each timestep (with reward $-0.1$), and also has a ``done'' action that it chooses to terminate at any position (with reward $2$ if it is a task goal). 
We train an agent on the task where it must learn to navigate to the middle of the top-left or bottom-right rooms. 
Figure~\ref{fig:WVF} (left) shows the learned WVF. The figure is generated by plotting the value functions for every goal position (since $\goals=\state$), and displaying it at their respective $xy$ position. Note how the values with respect the ``top-left'' and ``bottom-right'' goals are high (red), reflecting the high rewards the agent received for reaching those goals when it was trying to reach them.
Figure~\ref{fig:WVF} (middle) shows a close up view of the learned WVF around the ``top-left'' goal. 
We can observe from the value gradient of the plots that the WVF does indeed learn how to reach all positions in the gridworld.
We can then maximise over goals to obtain the regular value function and policy (Figure~\ref{fig:WVF} (right)). 

Finally, we demonstrate that the transition probabilities can be inferred from the learned WVF. 
Figures~\ref{fig:transitions} (left) and (middle) respectively show the transitions inferred by solving the Bellman equations with $s^\prime,g\in\state\times\state$ and $s^\prime,g\in\mathcal{N}(s)\times\mathcal{N}(s)$.
For each, we inferred the next state probabilities for taking each cardinal action at the center of each room, and placed the corresponding arrow in the state with highest probability.
The red arrows in Figure~\ref{fig:transitions} (left) correspond to incorrectly inferred next states, which is a consequence of the learned WVF not being near optimal at all states for all goals.  
Figure~\ref{fig:transitions} (middle) shows that in practice, if the WVF is not near-optimal, we can still infer dynamics by using $s^\prime,g\in\mathcal{N}(s)\times\mathcal{N}(s)$.
Figure~\ref{fig:transitions} (right) shows sample trajectories for following the optimal policy using the inferred transition probabilities. The gray-scale color of each arrow corresponds to the normalised value prediction for that state.

\begin{figure*}[h!]
\centering
\begin{minipage}{.48\textwidth}
    \centering
    \includegraphics[height=1.1in]{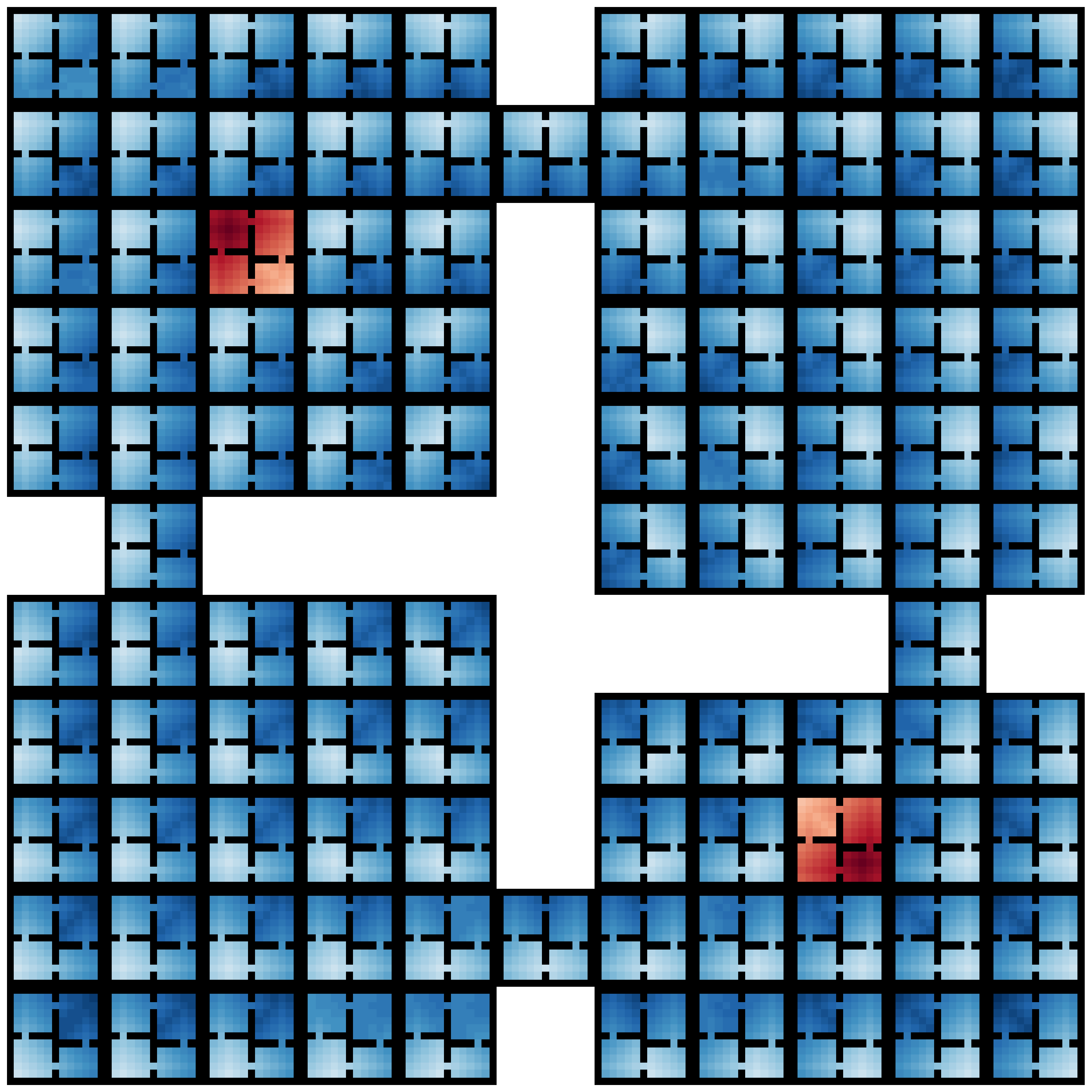}
    \includegraphics[height=1.1in]{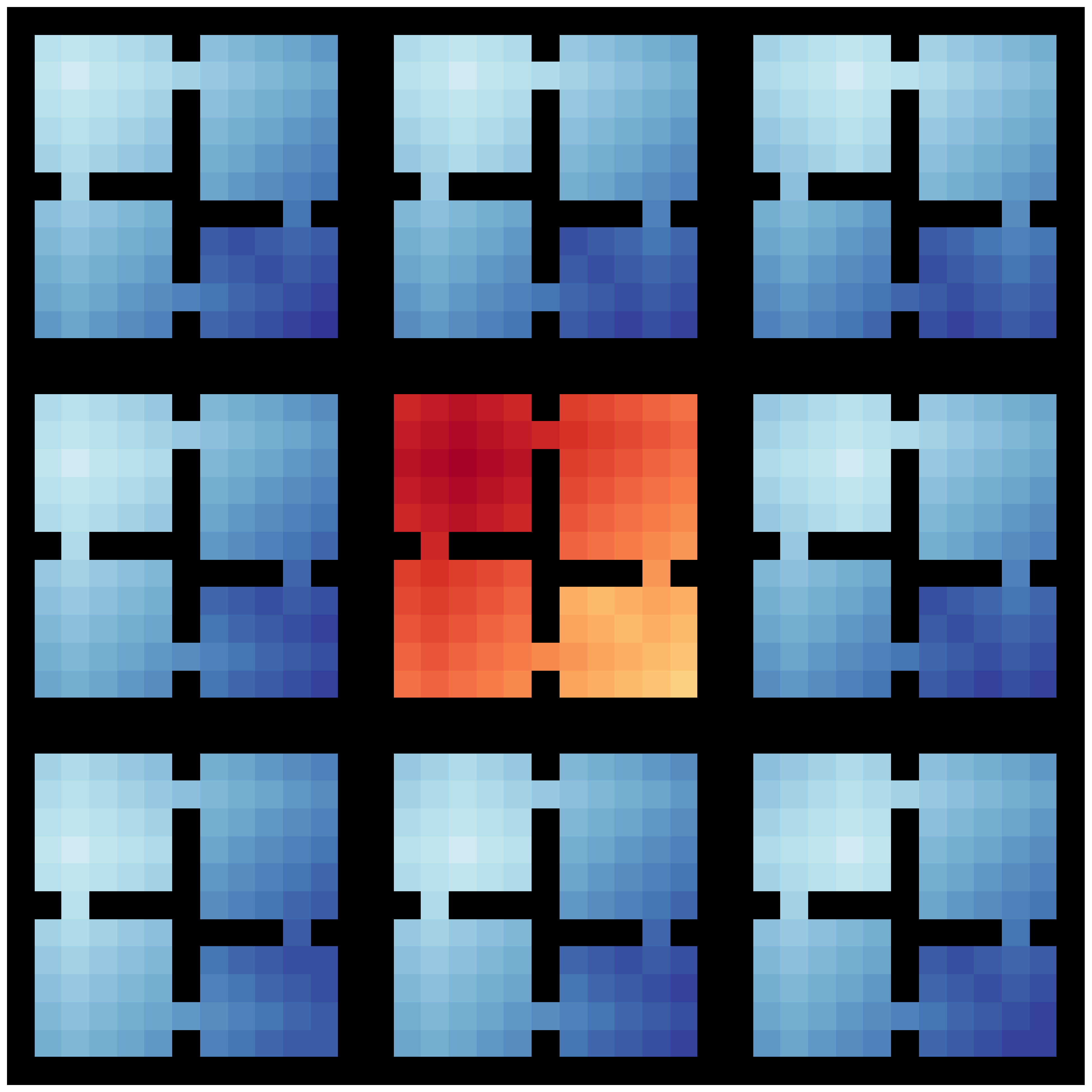}
    \includegraphics[height=1.1in]{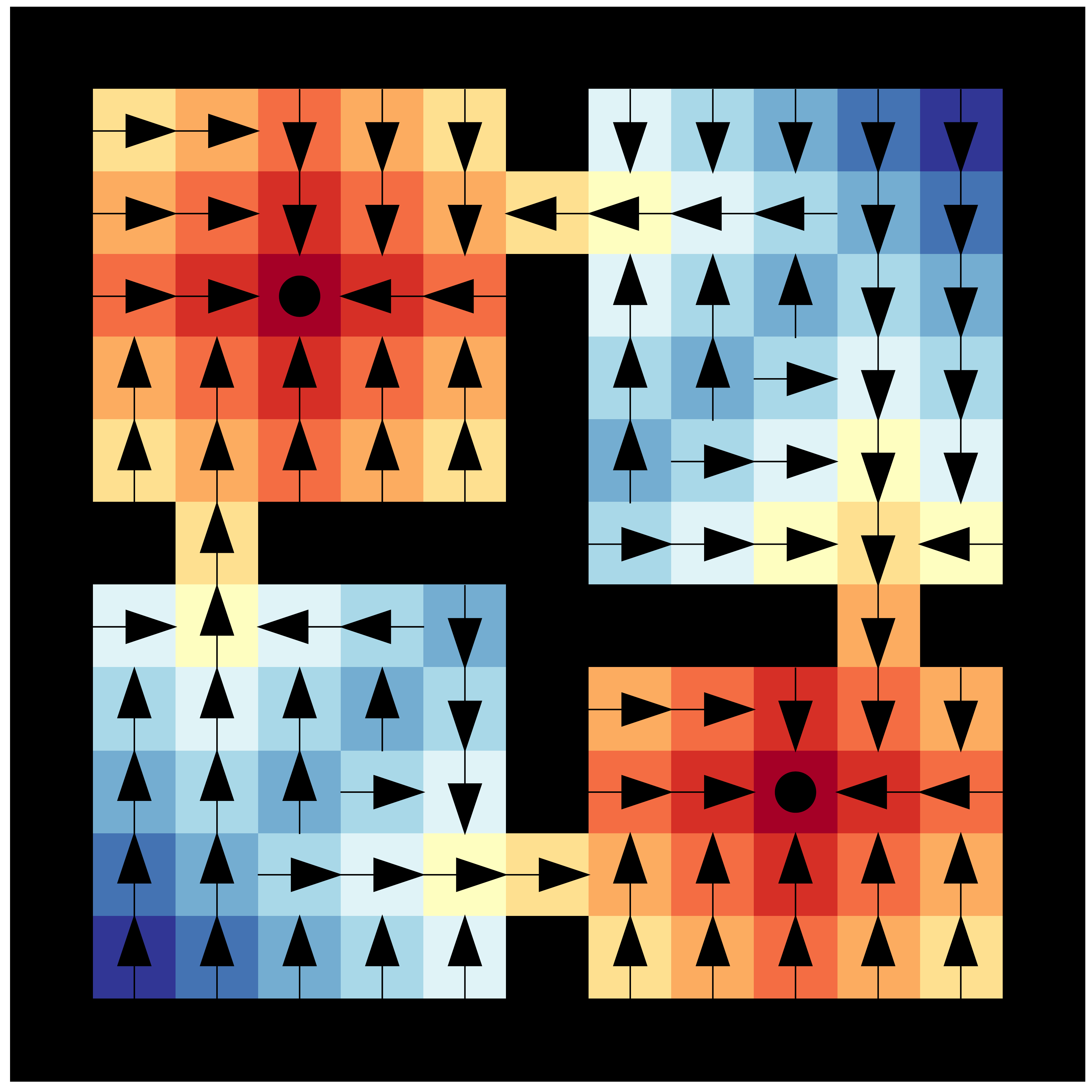}
    \captionof{figure}{Learned WVF (left), close up of WVF for ``top-left'' goal (middle), and inferred values and policy (right).}
    \label{fig:WVF}
\end{minipage}%
\quad
\begin{minipage}{.48\textwidth}
    \centering
    \includegraphics[width=1.1in]{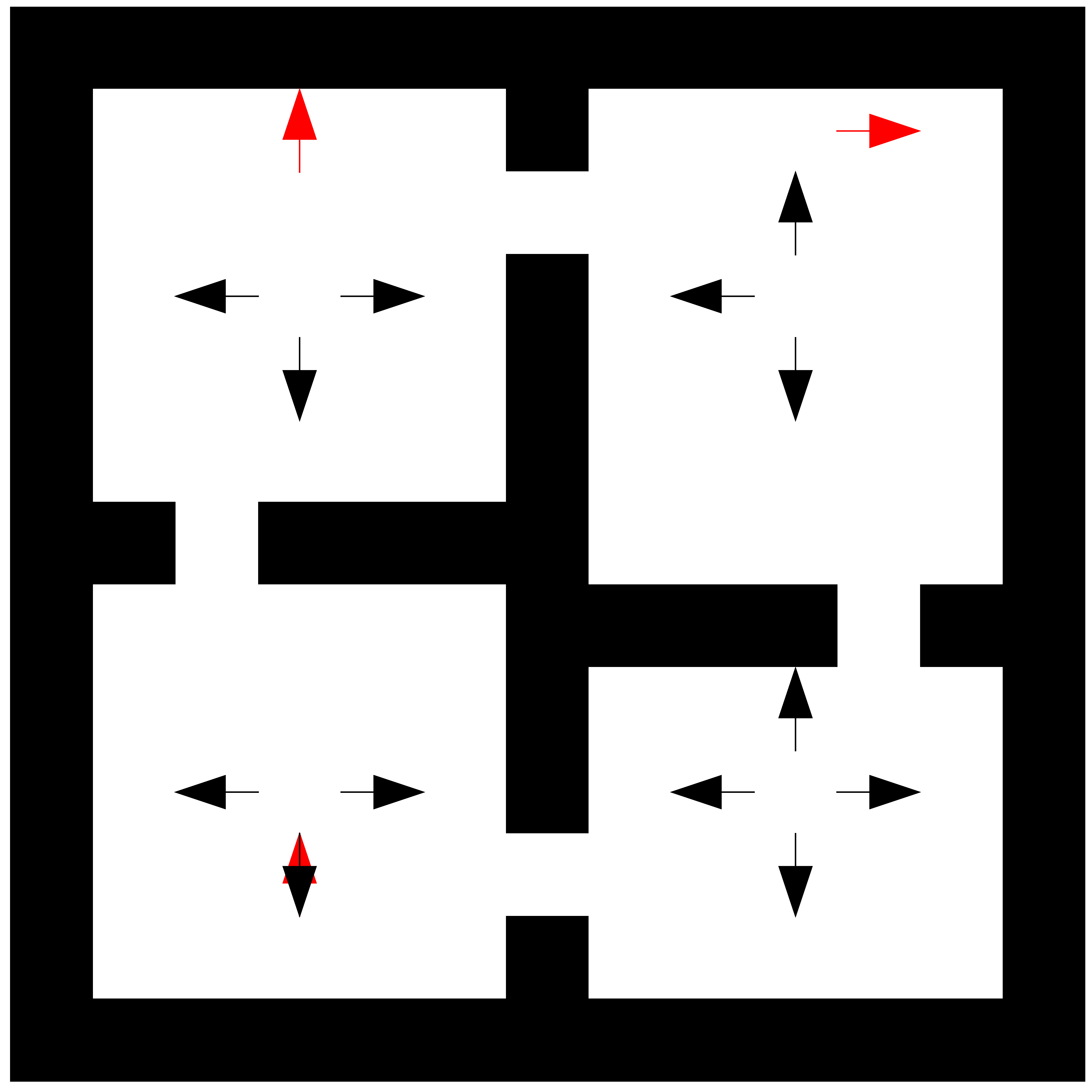}
    \includegraphics[width=1.1in]{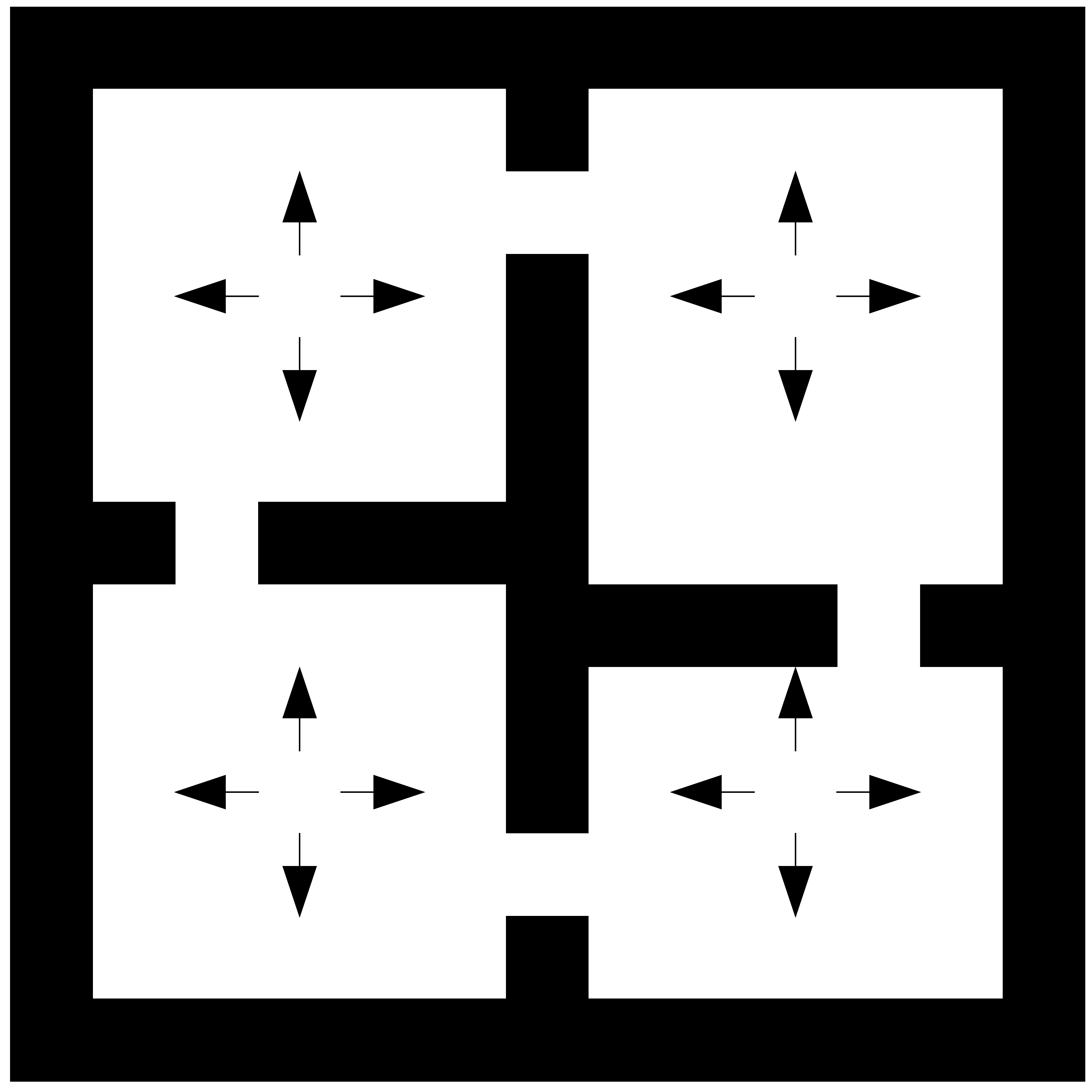}
    \includegraphics[width=1.1in]{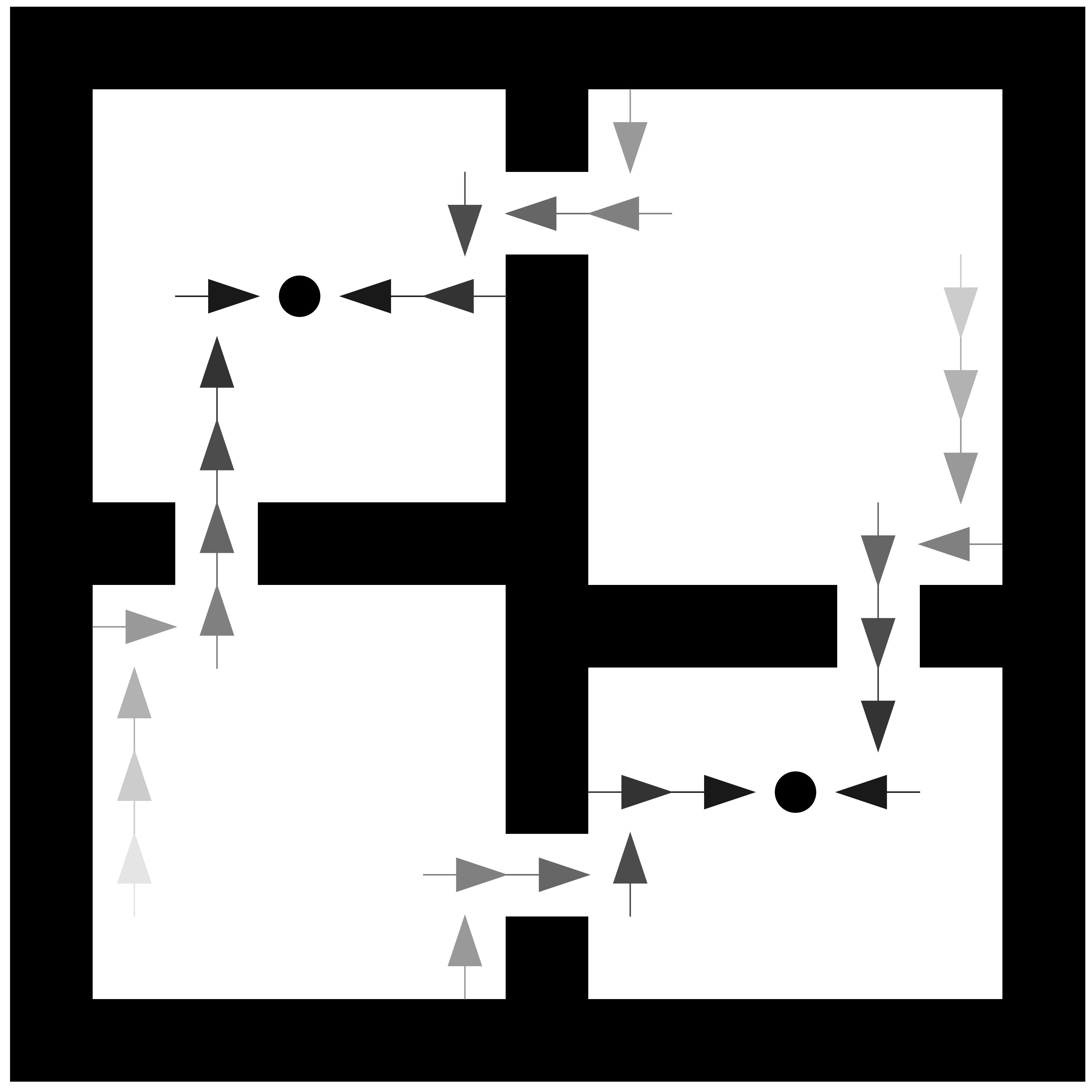}
    \captionof{figure}{Inferred transitions (left-middle) and imagined rollouts using the learned WVF (right).}
    \label{fig:transitions}
\end{minipage}
\end{figure*}

\end{experiment}

\section{Benefits of WVFs across tasks in the same world}
\label{sec:benefits}

We now highlight some benefits of WVFs that are a natural result of assuming tasks come from the same world---that is we assume that all tasks share the same state space, action space and dynamics, but differ in their reward functions. More specifically, we define the world as a background MDP $M_0 = \langle \mathcal{\state}_0, \action_0, \dynamics_0, \reward_0 \rangle$ with its own state space, action space, transition dynamics and background reward function. 
Any individual task $M$ is specified by a task-specific reward function $\reward_M^\tau(s,a)$ that is non-zero only for transitions entering terminal states.
The reward function for the resulting MDP is then simply $\reward_M(s,a,s^\prime) \coloneqq \reward_0(s,a,s^\prime) + \reward_M^\tau(s,a)$. We denote the set of all these tasks as $\tasks$ and the set of their corresponding optimal WVFs as $\goalq$.

Interestingly, this definition means that if tasks come from the same world, then their WVFs share the same world policy---that is, the agent has the same notion of goals and how to reach them no matter the task.
\begin{theorem} 
Let $\goalq$ be the set of optimal world $\bar{Q}$-value functions with mastery of tasks in $\tasks$. Then for all $s \neq g \in \state\times\goals$,
\[
\pibarstar(s,g) \in \argmax\limits_{a \in \action}\qstarbar_{M_1}(s, g, a) \iff \pibarstar(s,g) \in \argmax\limits_{a \in \action}\qstarbar_{M_2}(s, g, a) ~ \forall M_1, M_2 \in \tasks.
\]
\label{lem:pi1_e_pi2}
\end{theorem}
Similarly, if we require that the world policies need to be the same across tasks, then we have that the tasks must come from the same world as defined above.
In a sense, this gives us a notion of task invariance (the world does not change with changes in task) implying policy conservation (the world policy is constant across tasks), and vice-versa.

\subsection{Zero-shot values and policies from rewards}
\label{sec:R_WVF}

Since each task $M\in\tasks$ share the same dynamics (and consequently the same world policy), their corresponding WVFs can be written as $\qstarbar_M(s, g, a) = \gstar + \rbar_M^\tau(s^\prime, a^\prime)$ for some $s^\prime,a^\prime \in \state\times\action$, where $\gstar$ is a constant across tasks that represents the sum of rewards starting from $s$ and taking action $a$ up until $g$, but not including the terminal reward. Using this fact, we can show that the optimal value function and policy for any task can be obtained zero-shot from an arbitrary WVF given the task-specific rewards:

\begin{theorem}
Let $\reward_M^\tau$ be the given task-specific reward function for a task $M\in\tasks$, and let $\qstarbar\in\goalq$ be an arbitrary WVF. Define
\[
\vstarbara_M(s,g) := \max\limits_{a\in\action}\qstarbar(s,g,a) + \left(\max\limits_{a\in\action}\reward_M^\tau(g,a)-\max\limits_{a\in\action}\qstarbar(g,g,a)\right), \text{ the estimated WVF of } M.
\]
Then,
\begin{enumerate}[label=(\roman*)]
    \item for all $g\in\goals$ reachable from $s\in\state$, $\vstarbar_M(s,g) = \vstarbara_M(s,g)$.
    \item $\vstar_M(s) = \max\limits_{g\in\goals} \vstarbara(s,g)$, and $\pistar_M(s) \in \argmax\limits_{a\in\action}\qstarbar(s,\argmax\limits_{g\in\goals} \vstarbara_M(s,g),a)$.
\end{enumerate}

\label{thm:R_WVF}
\end{theorem}
This has several important implications for transfer learning. For example, one can learn an arbitrary WVF with unsupervised pretraining, then solve any new task by only learning the reward function (from experience or demonstrations).
\begin{experiment}

Consider the 4-rooms domain introduced in Experiment~\ref{exp:WVF} where the agent has learned the WVF for navigating to the top-left or bottom-right rooms.
Figure~\ref{fig:R_WVF} shows the estimated WVFs, values and policy for each task, computed as per Theorem~\ref{thm:R_WVF} using the given task rewards and the learned WVF for the bottom-left task.
Notice how the WVFs look similar to the rewards. This highlights the structural similarity between the task space and value function space which enables the zero-shot inference of task policies from task rewards alone.

\begin{figure*}[h!]
\centering
\begin{subfigure}{.5\textwidth}
    \centering
    \includegraphics[height=1.1in]{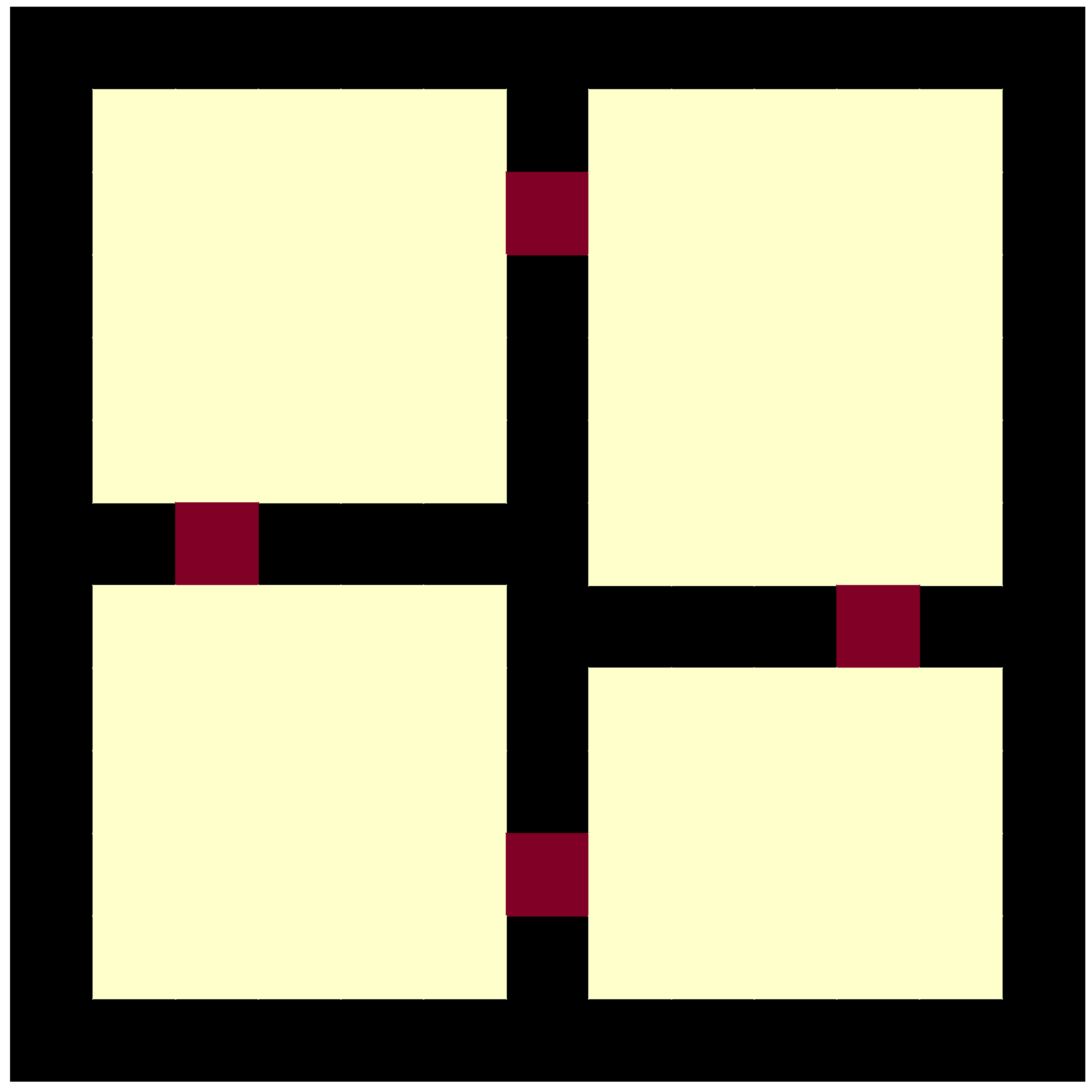}
    \includegraphics[height=1.1in]{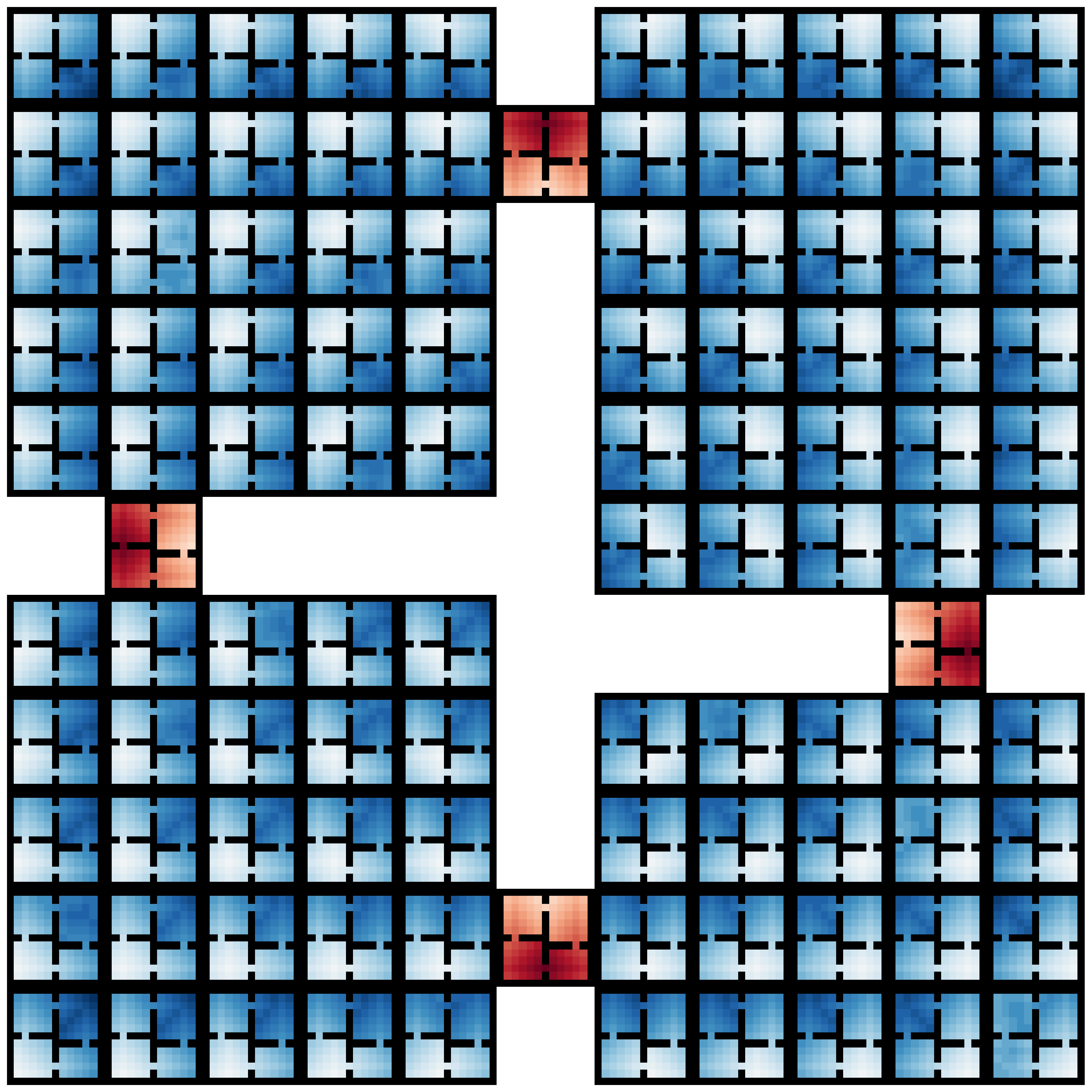}
    \includegraphics[height=1.1in]{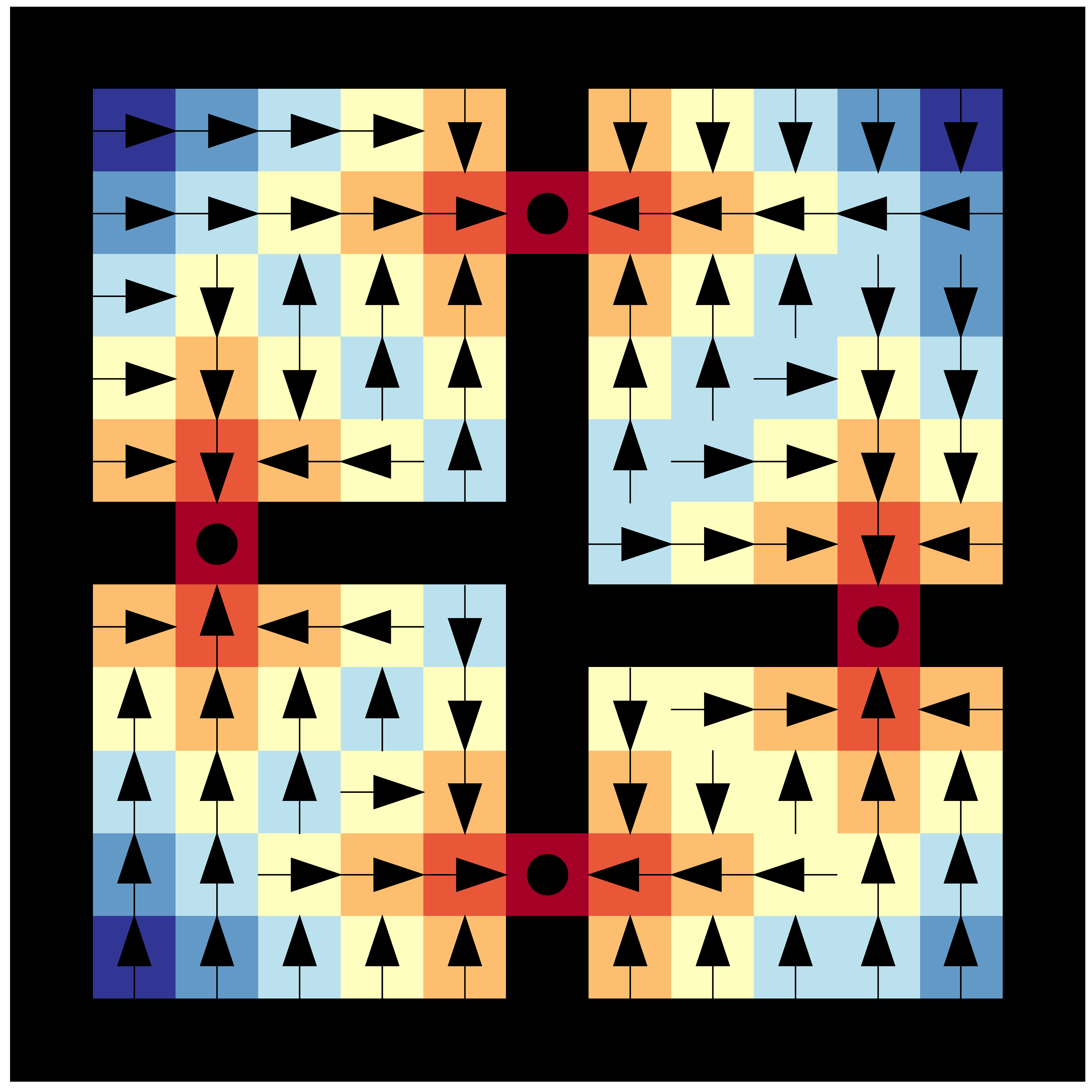}
    \caption{Navigating to the hallways.}
    \label{fig:task1}
\end{subfigure}%
\begin{subfigure}{.5\textwidth}
    \centering
    \includegraphics[height=1.1in]{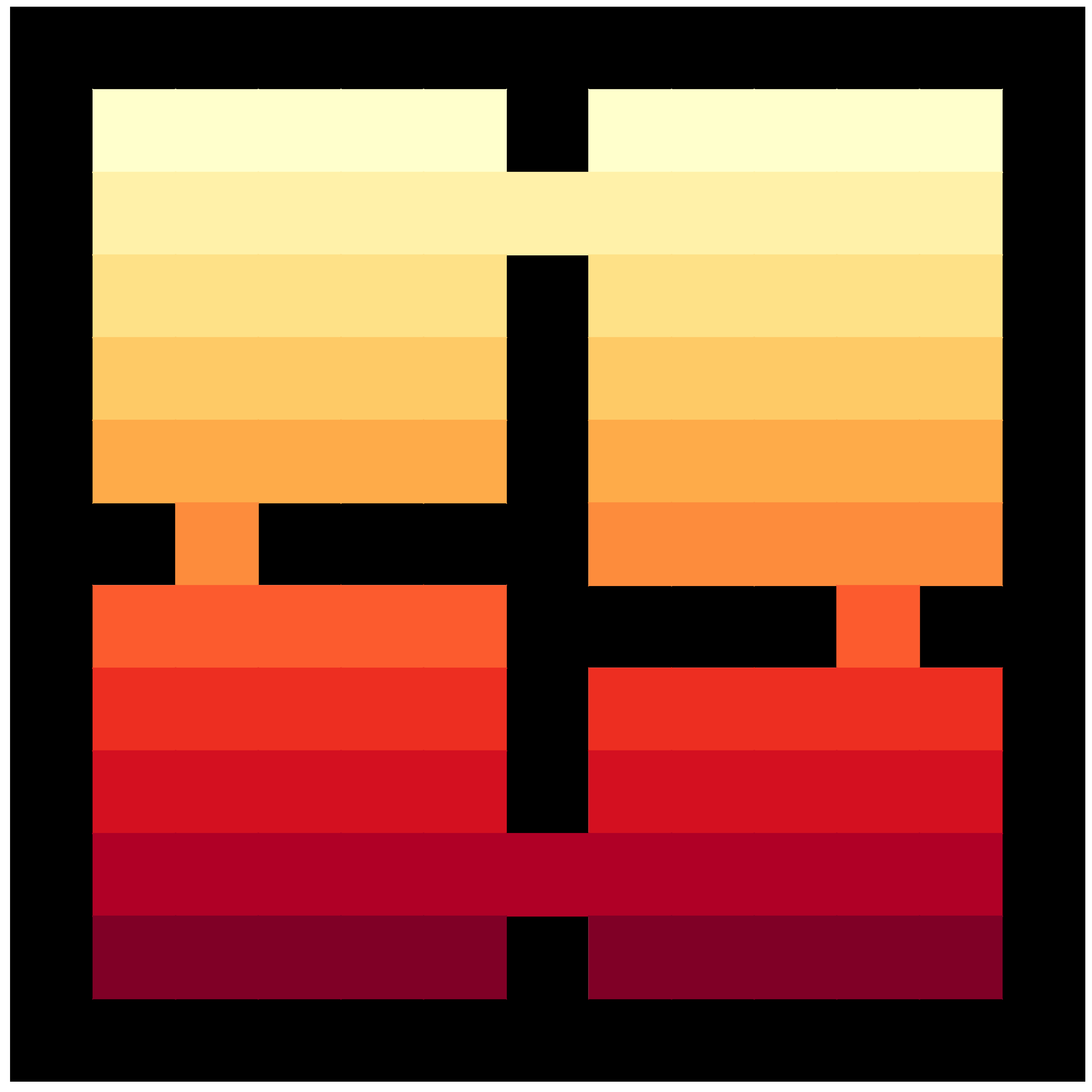}
    \includegraphics[height=1.1in]{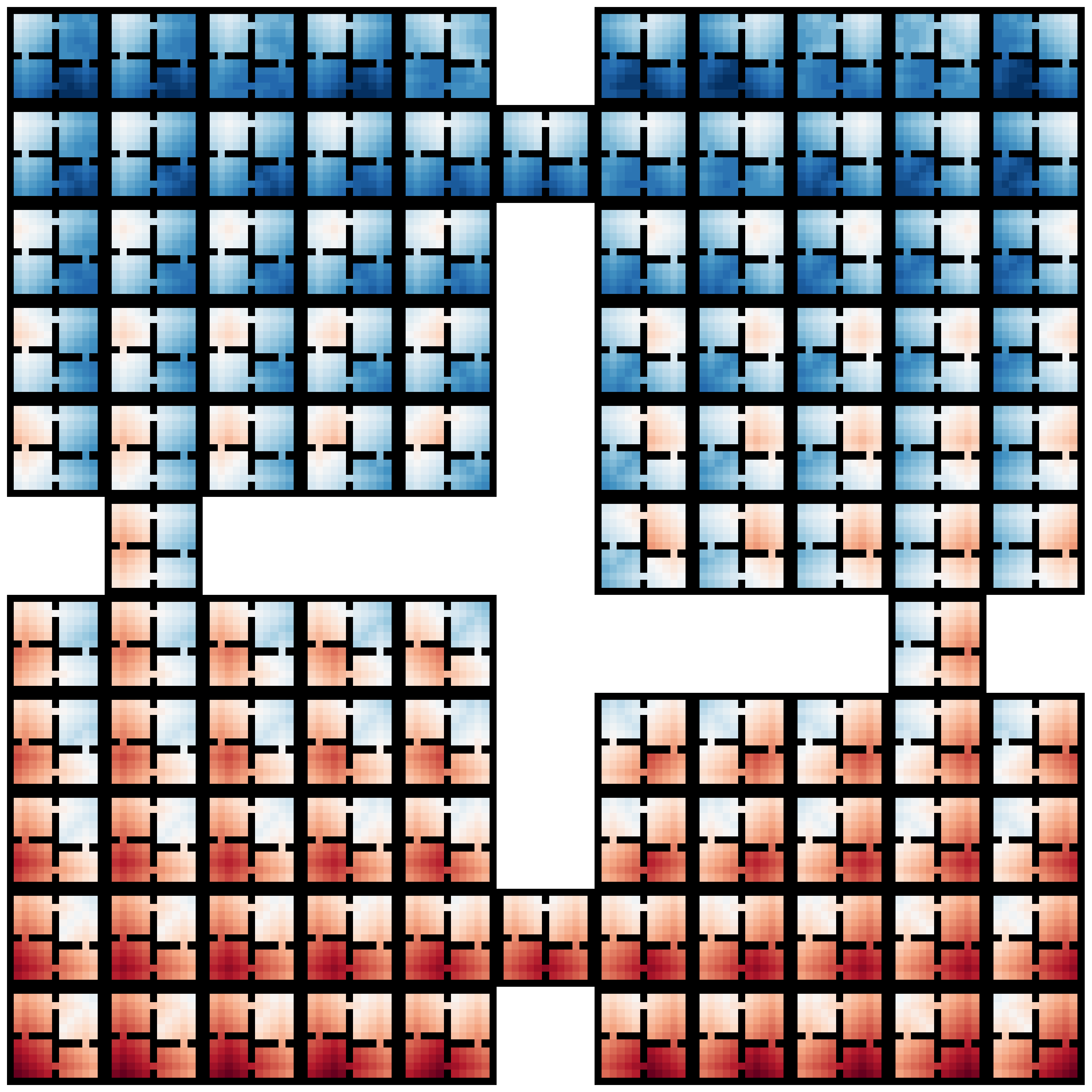}
    \includegraphics[height=1.1in]{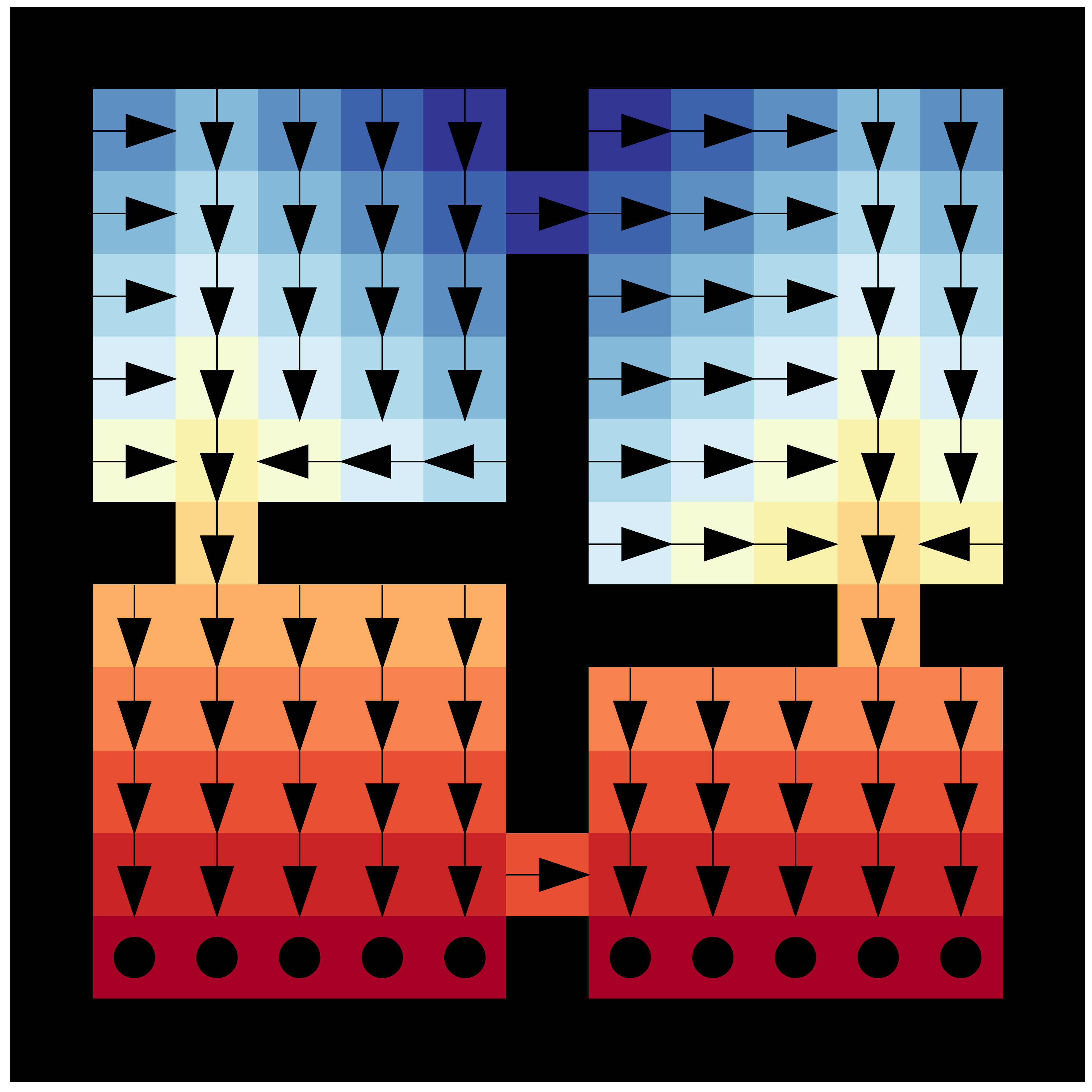}
    \caption{Navigating to the bottom of the grid.}
    \label{fig:task2}
\end{subfigure}%
\caption{Inferring values and policies zero-shot from goal rewards. From left to right on each figure: The given task specific rewards, the inferred WVF using Theorem~\ref{thm:R_WVF}, and the inferred values and policy from maximising over goals.}
\label{fig:R_WVF}
\end{figure*}

\end{experiment}

\subsection{Zero-shot logical composition}
\label{sec:logics}

We recently proposed a framework for agents to apply logical operations---conjunction ($\wedge$), disjunction ($\vee$) and negation ($\neg$)---over the space of tasks and WVFs \cite{nangue2020boolean}.
We first define these logical operations over tasks and WVFs as follows (we omit the value functions' parameters for readability):
\[
  \reward_{M_1 \vee M_2} \coloneqq  \max\{\reward_{M_1}, \reward_{M_2}\}
; \quad
  \reward_{M_1 \wedge M_2} \coloneqq \min\{\reward_{M_1}, \reward_{M_2}\}
; \quad
  \reward_{\neg M_1} \coloneqq \left(\reward_{\mbig} + \reward_{\msmall} \right) - \reward_{M_1},
\]
\[
  \qstarbar_{M_1} \vee \qstarbar_{M_2} \coloneqq  \max\{\qstarbar_{M_1}, \qstarbar_{M_2}\}
; \quad
  \qstarbar_{M_1} \wedge \qstarbar_{M_2} \coloneqq \min\{\qstarbar_{M_1}, \qstarbar_{M_2}\}
; \quad
  \neg \qstarbar_{M_1} \coloneqq \left(\qstarbarbig + \qstarbarsmall \right) - \qstarbar_{M_1},
\]
where $\qstarbarbig$ is the goal-oriented value function for the maximum task $\reward_{\mbig}$ where $r = \rmax$ for all $\goals$.
Similarly $\qstarbarsmall$ is goal-oriented value function  for the minimum task $\reward_{\msmall}$ where $r = \rmin$ for all $\goals$. 

We then showed that the logical composition of WVFs produces exactly the WVF for the corresponding task composition (Theorem~\ref{thm:composition} (i)).  
We further showed that when the task specific rewards are binary, $\tasks$ and $\goalq$ from homomorphic Boolean algebras (Theorem~\ref{thm:composition} (ii)). This formally defines classical logic over the space of tasks and value functions.

\begin{theorem}
Let $\goalq$ be the set of optimal world $\bar{Q}$-value functions for tasks in $\tasks$. Then, 
\begin{enumerate}[label=(\roman*)]
    \item for all tasks $M_1, M_2 \in \tasks$, we have $\qstarbar_{\neg M_1} = \neg \qstarbar_{M_1}$, $\qstarbar_{M_1 \vee M_2} = \qstarbar_{M_1} \vee \qstarbar_{M_2}$, and  $\qstarbar_{M_1 \wedge M_2} = \qstarbar_{M_1} \wedge \qstarbar_{M_2}$.
    \item If the task specific rewards are binary, then $(\tasks,\vee,\wedge,\neg)$ and $(\goalq,\vee,\wedge,\neg)$ are homomorphic Boolean algebras.
\end{enumerate}

\label{thm:composition}
\end{theorem}

\begin{experiment}

Consider video game world where an agent must navigate a 2D world and collect objects of different shapes and colours \cite{vanniekerk19}.
The state space is an $84 \times 84$ RGB image, and the agent is able to move in any of the four cardinal directions.
The agent also possesses a \texttt{pick-up} action, which allows it to collect an object when standing on top of it.
The positions of the agent and objects are randomised at the start of each episode.     
We first use deep Q-learning (similarly to Algorithm~\ref{algo:qlearn}) to learn two base tasks: collecting blue objects and collecting squares.
Figure~\ref{fig:WVF_Blue} shows the values with respect to each goal of the learned WVF for collecting blue objects, and maximising over goals gives us the regular value function (Figure~\ref{fig:VF_Blue}).
%
We can then compose them to solve their disjunction (\textit{OR}), conjunction (\textit{AND}) and exlusive-or (\textit{XOR}), as shown in Figure~\ref{fig:returns}. In general, after learning $n$ base tasks we can solve all their $2^{2^n}$ logical compositions, giving agents a super-exponential explosion of skills (Figure~\ref{fig:analytic}).

\begin{figure*}[h!]
    \centering
    \begin{subfigure}[t]{0.2\textwidth}
        \centering
        \includegraphics[height=1.3in]{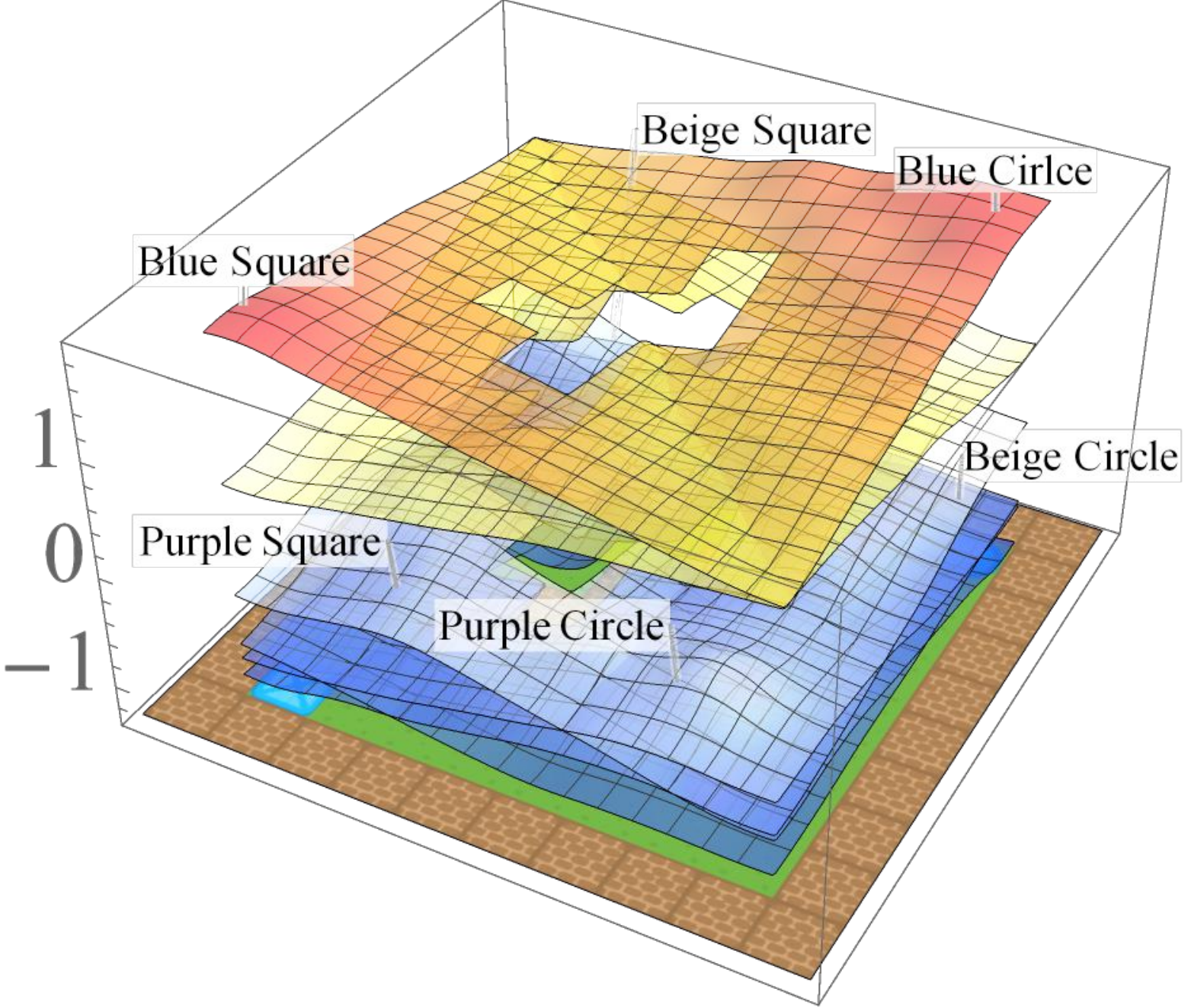}
        \caption[caption]{Learned WVF for collecting blue objects.}
         \label{fig:WVF_Blue}
    \end{subfigure}%
    ~
    \begin{subfigure}[t]{0.22\textwidth}
        \centering
        \includegraphics[height=1.1in]{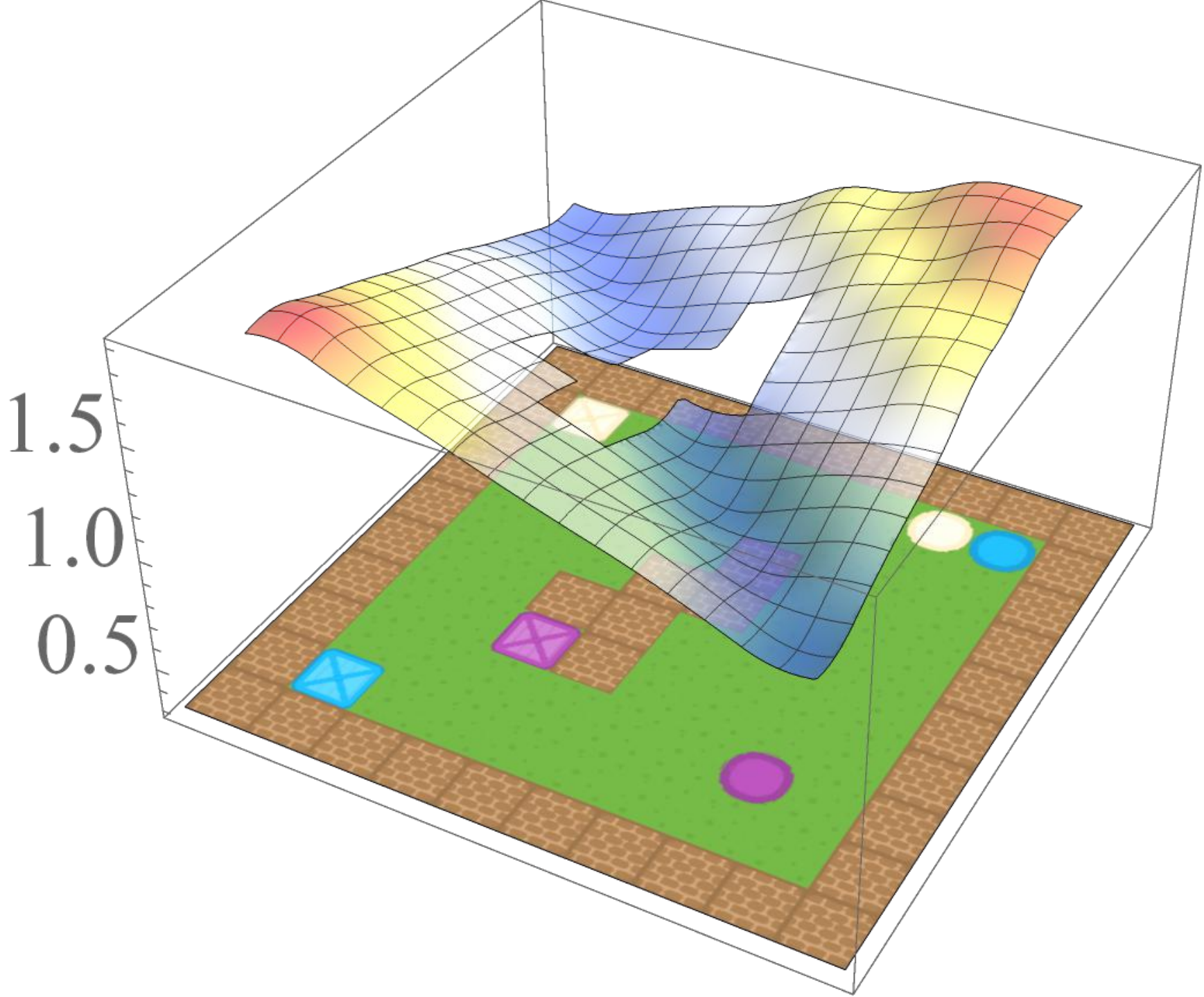}
        \caption[caption]{Inferred regular value function for collecting $Blue$ objects.}
         \label{fig:VF_Blue}
    \end{subfigure}%
    ~
    \begin{subfigure}[t]{0.3\textwidth}
        \centering
        \includegraphics[height=1.1in]{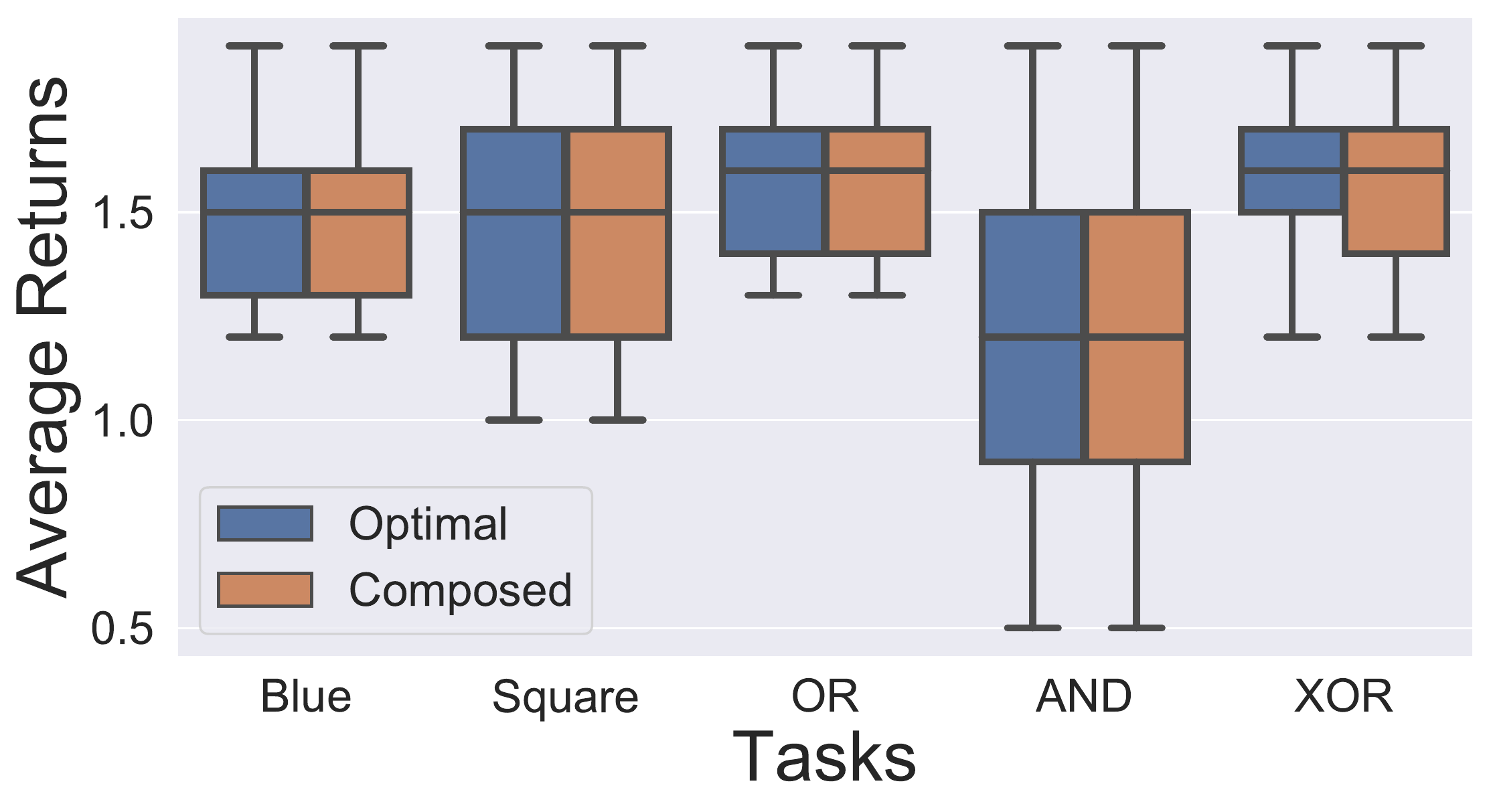}
        \caption[caption]{Average returns over 1000 episodes of optimal and composed WVFs.}
         \label{fig:returns}
    \end{subfigure}%
    ~
    \begin{subfigure}[t]{0.2\textwidth}
        \centering
        \includegraphics[height=1.3in]{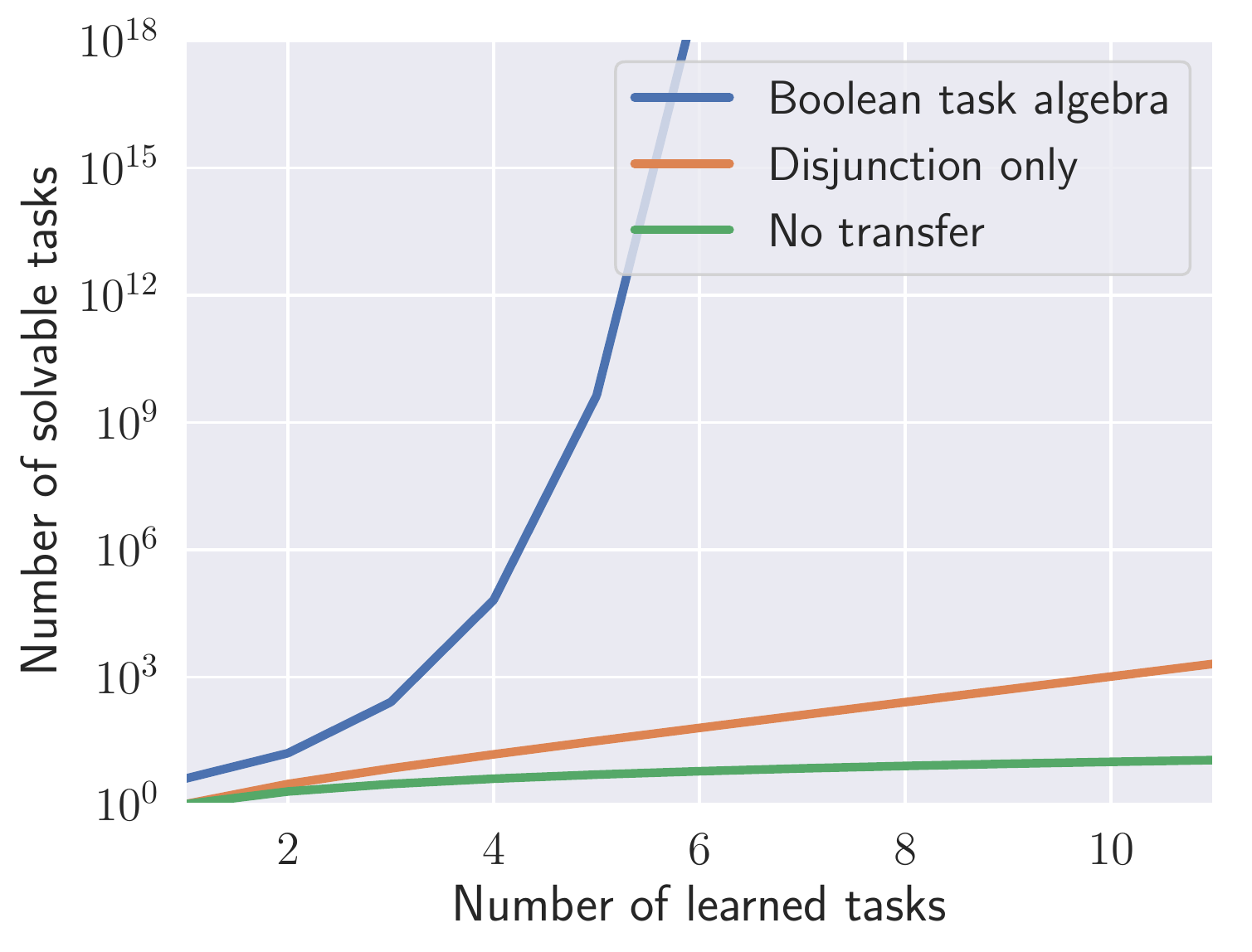}
        \caption[caption]{Super-exponential number of solvable tasks from learned ones.}
         \label{fig:analytic}
    \end{subfigure}%
    \caption{ Experiments in a 2D domain where the agent needs to collect objects of various shapes and colors.  }
    \label{fig:composition}
\end{figure*}


\end{experiment}

\section{Conclusion}

For an agent to learn how to solve multiple tasks in a given world from a single stream of experience, it should not only encode knowledge about how to optimally solve the current task, but also how achieve agent-centric goals in the world. 
We introduced WVFs as a principled knowledge representation that captures this idea. We then showed that WVFs learned on any task have mastery of the world---they encode how to solve all possible goal-reaching tasks. This results in agents that can infer the dynamics of the environment (which can be used for model based approaches like planning), enables them to solve new tasks zero-shot given just their terminal rewards, and finally enables zero-shot logical composition of learned skills (an important ability for the combinatorial explosion of skills in long-lived agents).

{\footnotesize
\bibliography{rldm}
\bibliographystyle{plain}
}
\end{document}